\documentclass[10pt,twocolumn,letterpaper]{article}

\usepackage[pagenumbers]{cvpr}

\usepackage{microtype}
\renewcommand{\paragraph}[1]{\vspace{.2em}\noindent\textbf{#1}}
\DeclareMathOperator*{\argmin}{arg\,min}
\DeclareMathOperator*{\argmax}{arg\,max}

\usepackage{multirow}
\usepackage{colortbl}
\usepackage{tabularx}
\usepackage{algorithm}
\usepackage{algpseudocode}
\usepackage{pifont}
\newcommand{\cmark}{\ding{51}}
\newcommand{\xmark}{\ding{55}}

\newcommand{\method}{DPR\xspace}

\usepackage{tikz}
\usetikzlibrary{positioning,arrows.meta,fit,calc,backgrounds,decorations.pathreplacing,shapes.geometric}

\definecolor{bpOurs}{RGB}{0,114,178}      
\definecolor{bpPrior}{RGB}{213,94,0}      
\definecolor{bpData}{RGB}{0,158,115}      
\definecolor{bpAccent}{RGB}{230,159,0}    
\definecolor{bpGray}{RGB}{110,110,110}    
\definecolor{bpLight}{RGB}{238,238,238}   
\colorlet{bpOursFill}{bpOurs!18}
\colorlet{bpPriorFill}{bpPrior!15}
\colorlet{bpDataFill}{bpData!15}

\tikzset{
  bp/base/.style   ={draw=black!70, line width=0.5pt, rounded corners=2pt, align=center,
                     font=\sffamily\scriptsize, minimum height=0.6cm, inner sep=3pt},
  bp/module/.style ={bp/base, fill=white},
  bp/frozen/.style ={bp/base, fill=bpLight, draw=bpGray, text=black!80},
  bp/ours/.style   ={bp/base, fill=bpOursFill, draw=bpOurs, line width=0.7pt},
  bp/prior/.style  ={bp/base, fill=bpPriorFill, draw=bpPrior},
  bp/data/.style   ={bp/base, fill=bpDataFill, draw=bpData},
  bp/todo/.style   ={bp/base, dashed, draw=bpGray, text=bpGray},
  bp/flow/.style   ={-{Latex[length=1.6mm]}, line width=0.6pt, black!75},
  bp/note/.style   ={font=\sffamily\tiny, text=bpGray, align=center},
  bp/panel/.style  ={font=\sffamily\footnotesize\bfseries, anchor=west},
  bp/frame/.style  ={draw=black!70, fill=white, line width=0.4pt},
}

\definecolor{cvprblue}{rgb}{0.21,0.49,0.74}
\usepackage[breaklinks,colorlinks,allcolors=cvprblue]{hyperref}

\def\confName{CVPR}
\def\confYear{2027}

\title{From Joint Risk to Decision Invariance: Decision-Preserving Regularization for Exemplar-Free Class-Incremental Learning}

\author{Qi Zhu \quad Ziang Gan \quad Wanting Zhang \quad Libao Zhang\thanks{Corresponding author: \href{mailto:libaozhang@bnu.edu.cn}{\texttt{libaozhang@bnu.edu.cn}}.}\\
School of Artificial Intelligence\\
Beijing Normal University, Beijing 100875, China
}

\hypersetup{
    pdftitle={From Joint Risk to Decision Invariance: Decision-Preserving Regularization for Exemplar-Free Class-Incremental Learning},
    pdfauthor={Qi Zhu, Ziang Gan, Wanting Zhang, Libao Zhang}
}

\begin{document}
\maketitle
\begin{abstract}
Exemplar-free continual learning seeks to minimize joint risk over all seen tasks, yet historical data are unavailable for direct optimization. We observe that, for classification, preserving the model's decisions on previously learned classes helps retain old knowledge. This motivates Decision-Preserving Regularization (\method). Penalizing changes in all pairwise old-class logit margins reduces algebraically to a centered relative-logit penalty. We further apply spectral whitening to remove the classifier's spectral scale bias. Because the global classifier supplies the decision constraints for the next stage, it must remain calibrated across all seen classes despite the drift of stored old-class statistics during representation updates. We therefore fit a closed-form affine map from paired current-task features to transport old-class Gaussian means and covariances into the updated feature space, then calibrate the global classifier using these distributions and current-task features. Across CIFAR-100, Tiny-ImageNet, ImageNet-Subset, and ImageNet-1K under cold-start protocols, comparisons with state-of-the-art methods show that \method achieves higher accuracy in most settings while remaining highly efficient. Code is available at \url{https://github.com/bnufw/dpr}.
\end{abstract}

\section{Introduction}
\label{sec:intro}

\begin{figure}[t]
\centering
\includegraphics[width=\linewidth]{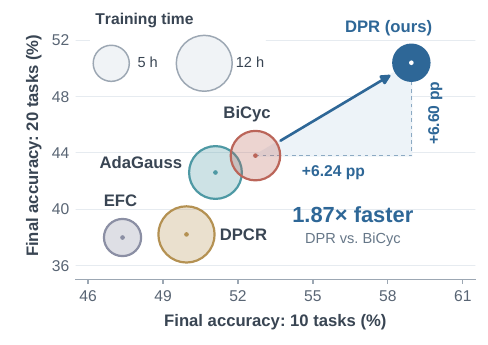}
\caption{Final accuracy on ImageNet-Subset. Bubble area indicates full-run wall-clock time for 10 tasks; smaller is faster. (Sec.~\ref{sec:efficiency}).}
\label{fig:teaser}
\end{figure}

Exemplar-free class-incremental learning (EFCIL) trains a single model on a sequence of disjoint class sets without retaining historical images, so that inference distinguishes all seen classes without a task identifier~\cite{delange2021continual,masana2023class}. The ideal goal of continual learning is to minimize the joint empirical risk over all observed tasks. Because historical data are absent, this joint objective cannot be optimized directly. Under the cold-start protocol, a small initial task provides an incomplete representation and the feature extractor must adapt substantially in subsequent stages to separate new classes~\cite{magistri2024elastic}. Updating the backbone on current data alone shifts old-class features across established decision boundaries and causes catastrophic forgetting~\cite{mccloskey1989catastrophic,french1999catastrophic}.

Existing EFCIL methods approximate old-task retention through surrogate feature regularizers and post-hoc drift compensators, yet both designs remain indirect proxies for the old model's classification decisions. Isotropic feature distillation penalizes all feature displacements equally and restricts new-class plasticity. Elastic Feature Consolidation (EFC)~\cite{magistri2024elastic} estimates an Empirical Feature Matrix over training samples to constrain sensitive directions, while projection methods partition feature or gradient updates~\cite{wang2023orthogonal,zhao2023rethinking,he2025semantic}. However, data-dependent sensitivity matrices require sample-level curvature estimation and dense matrix storage without directly enforcing decision invariance. Meanwhile, as the backbone evolves on new classes, stored old-class means and covariances drift out of alignment with the updated feature space. Prior methods compensate for this shift by training auxiliary mapping networks across dozens of epochs~\cite{gomez2024exemplar,xu2026two} or optimizing multi-step adversarial perturbations through the backbone~\cite{goswami2024resurrecting}. Can we instead reformulate the intractable joint-learning objective directly into a tractable classification target?

For classification, retaining old knowledge does not require reproducing historical risk or freezing the feature vector numerically. It requires only that the model's old-class classification decisions remain invariant under the frozen old classifier head. Translating this principle into a continuous objective requires preserving the relative ordering of old-class logits, which we formulate by keeping all pairwise logit margins between old classes invariant. By algebraic identity, penalizing all pairwise margin drifts over $K$ old classes collapses into a single quadratic norm on the row-centered old classifier head $W_c$, allowing arbitrary shared logit offsets while preserving both $\argmax$ predictions and conditional softmax probabilities. However, directly penalizing centered logit drift weights each discriminative direction by the squared spectral scale of $W_c$. We introduce \emph{Decision-Preserving Regularization} (\method), which applies spectral whitening to the centered classifier head to decouple spectral scale bias and extract an orthonormal discriminative basis $V \in \mathbb{R}^{d \times r}$, constraining all $r$ active decision directions on an equal footing while applying a mild smoothness penalty to the complementary space for new-class learning.

Because \method extracts its discriminative constraints directly from the old classifier head, sustaining decision preservation across stages relies on a well-calibrated global classifier over all seen classes; yet representation updates on new classes train only a local task head and cause stored old-class distributions to drift. Rather than training iterative drift networks as in LDC~\cite{gomez2024exemplar} and BiCyc~\cite{xu2026two} or backpropagating adversarial perturbations as in ADC~\cite{goswami2024resurrecting}, we extract paired features from the frozen old and updated backbones in a single forward pass over current data and solve ridge regression with an identity prior in closed form for a shared affine map. This affine transformation transports both stored old-class means and covariances into the updated representation space in milliseconds. We then sample synthetic old-class features from the transported Gaussian distributions alongside real current-task features to calibrate a single bias-free linear classifier head directly in feature space. Unlike Mahalanobis classifiers such as AdaGauss~\cite{rypesc2024task} and BiCyc~\cite{xu2026two}, our calibrated linear head requires no extra projection layers and allows all stored class statistics to be discarded after incremental training.

Because the derived objective reduces to a single low-rank matrix product per sample during training and a closed-form linear solve afterward, \method is simple, efficient, and effective (Fig.~\ref{fig:teaser}). Across CIFAR-100~\cite{krizhevsky2009learning}, Tiny-ImageNet~\cite{wu2017tiny}, ImageNet-Subset~\cite{deng2009imagenet,douillard2020podnet}, and large-scale ImageNet-1K~\cite{deng2009imagenet} under ten- and twenty-task cold-start protocols, \method outperforms prior EFCIL baselines such as BiCyc and DPCR. On ImageNet-Subset, \method improves final accuracy over BiCyc by 6.24 and 6.60 pp on the two task splits, and it scales to ImageNet-1K with gains of up to 3.83 pp over BiCyc. Component-wise profiling on Tiny-ImageNet confirms that spectral whitening, decision-preserving regularization, closed-form affine transport, and classifier calibration together add only 1.39\% to per-task training time, completing ten-task ImageNet-Subset training in 5.04 hours (1.87$\times$ faster than BiCyc). Formulating continual classification directly from decision invariance thus yields a simple and practical alternative to empirical sensitivity matrices and auxiliary drift networks.

In summary, our contributions are threefold:
\begin{itemize}
    \item We reformulate the joint-learning objective as new-class learning with old-class decision preservation and derive a tractable relative-logit regularizer.
    \item We propose \method, combining spectral whitening for decision preservation with closed-form affine compensation for global classifier calibration.
    \item Experiments on four cold-start benchmarks demonstrate improved accuracy and a better balance between old-class retention and new-class learning, with low computational overhead.
\end{itemize}

\begin{figure*}[t]
\centering
\includegraphics[width=\textwidth]{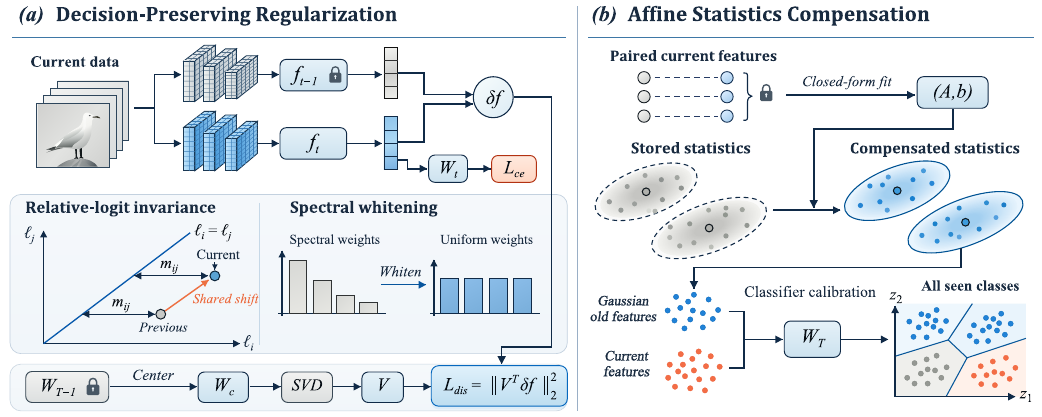}
\caption{\textbf{Overview of \method.}
(a) Decision-preserving regularization preserves pairwise old-class margins while allowing a shared logit shift. Spectral whitening of the centered old classifier yields a basis $V$ for uniformly penalizing feature drift along old-class discriminative directions during new-class learning.
(b) A closed-form affine map fitted to paired current-task features compensates stored old-class means and covariances for representation drift. Gaussian old-class features sampled from the compensated statistics are combined with real current-task features to calibrate the global head $W_T$ for unified inference and next-stage decision regularization.}
\label{fig:overview}
\end{figure*}

\section{Related Work}
\label{sec:related}

\paragraph{Continual learning and directional constraints.}
Continual learning methods preserve previously acquired knowledge through parameter regularization, knowledge distillation, replay, or architectural expansion~\cite{delange2021continual,masana2023class}. Parameter-level regularizers such as EWC~\cite{kirkpatrick2017overcoming}, SI~\cite{zenke2017continual}, and MAS~\cite{aljundi2018memory} estimate weight importance and penalize deviations from old parameters, while distillation methods constrain output logits or intermediate activations~\cite{li2018learning,hou2019learning,douillard2020podnet}. To balance stability and plasticity more selectively, directional and gradient-projection methods partition the representation or gradient updates~\cite{wang2023orthogonal,zhao2023rethinking}. Rather than estimating empirical covariance or gradient directions from data, our approach reformulates old-task retention as classification-decision invariance and derives the active discriminative directions algebraically from pairwise logit-margin preservation under the frozen linear classifier head.

\paragraph{Exemplar-free representation regularization.}
When historical exemplars cannot be stored, representation learning must rely solely on current-task data and compact model states. PASS~\cite{zhu2021prototype} and SSRE~\cite{zhu2022self} combine prototype augmentation with self-supervised or structural constraints, often starting from a large initial task. Under the more demanding cold-start setting, EFC~\cite{magistri2024elastic} and EFC++~\cite{magistri2025efc} compute an Empirical Feature Matrix over training samples to regularize sensitive feature directions, and DPCR~\cite{he2025semantic} couples dual-projection constraints with classifier reconstruction. Whereas EFC requires $\mathcal{O}(Bd^2)$ per-batch operations and stores a dense $d\times d$ sensitivity matrix, \method applies spectral whitening to the row-centered old classifier head to constrain the $r$ active decision directions in $\mathcal{O}(Bdr)$ time.

\paragraph{Prototype drift compensation.}
Summarizing old classes via feature statistics enables decision boundary maintenance without raw images. FeTrIL~\cite{petit2023fetril} translates pseudo-features under a frozen backbone, and FeCAM~\cite{goswami2023fecam} exploits per-class covariance matrices for Mahalanobis classification. When the backbone adapts across tasks, stored prototypes and covariances suffer from representation drift. SDC~\cite{yu2020semantic} estimates semantic drift via kernel-weighted displacements, ADC~\cite{goswami2024resurrecting} synthesizes adversarial samples through the backbone to estimate drift, LDC~\cite{gomez2024exemplar} trains learnable projection networks across epochs, AdaGauss~\cite{rypesc2024task} adapts low-dimensional covariances to mitigate task-recency bias, and BiCyc~\cite{xu2026two} optimizes bidirectional alignment networks with cycle consistency. In contrast to iterative drift networks and Mahalanobis classifiers that retain auxiliary projection layers at inference, we solve a single ridge-regularized affine system in closed form to transport both old-class means and covariances.

\section{Method}
\label{sec:method}

We revisit the continual learning objective for classification, using the idealized joint-learning objective to motivate decision-preserving feature learning (Sec.~\ref{sec:problem}). We then derive a tractable centered relative-logit penalty from pairwise decision margins (Sec.~\ref{sec:logit_derivation}) and apply spectral whitening to remove spectral scale bias from the feature training loss (Sec.~\ref{sec:whitening}). Finally, because \method relies on a well-calibrated global classifier at each stage to anchor subsequent decision constraints, we compensate for the drift of stored old-class statistics via a closed-form affine transformation and calibrate the global classifier (Sec.~\ref{sec:affine_compensation}). Figure~\ref{fig:overview} provides an overview of \method.

\subsection{From Joint-Learning Upper Bound to Decision Preservation}
\label{sec:problem}

Consider a sequence of $N$ incremental tasks $\{\mathcal{D}_k\}_{k=1}^{N}$ with mutually disjoint class sets $\mathcal{C}_k$. At incremental stage $t$, the learner accesses only the current dataset $\mathcal{D}_t$ without retaining any historical exemplars. Let $\mathcal{C}_{\text{old}} = \bigcup_{k=1}^{t-1} \mathcal{C}_k$ denote the set of $K = |\mathcal{C}_{\text{old}}|$ previously seen classes, and $\mathcal{C}_{1:t} = \mathcal{C}_{\text{old}} \cup \mathcal{C}_t$ denote all classes observed up to stage $t$. We decouple the network into a feature extractor $f_{\theta_t}: \mathcal{X} \to \mathbb{R}^d$ and bias-free linear classification heads, where $W_t \in \mathbb{R}^{|\mathcal{C}_t| \times d}$ denotes the task-specific classifier head for task $t$, and $W_T \in \mathbb{R}^{|\mathcal{C}_{1:t}| \times d}$ denotes the global classifier head covering tasks $1$ through $t$.

The general goal of continual learning is to approximate the performance upper bound of joint training over all observed tasks~\cite{li2018learning,delange2021continual,masana2023class}:
\begin{equation}
(\theta_t^*, W_T^*) = \argmin_{\theta_t, W_T} \sum_{k=1}^{t} \mathbb{E}_{(\mathbf{x}, y) \sim \mathcal{D}_k} \left[ \mathcal{L}_{\text{ce}}\left( W_T f_{\theta_t}(\mathbf{x}), y \right) \right].
\label{eq:efcil_objective}
\end{equation}
However, because historical data $\{\mathcal{D}_k\}_{k=1}^{t-1}$ are unavailable, directly optimizing Eq.~\eqref{eq:efcil_objective} is intractable.

We notice that for classification problems, a sufficient condition of retaining old knowledge is to preserve the model's classification decisions on old classes. Let $f_{\theta_{t-1}}$ be the frozen feature extractor from stage $t-1$ and $W_{T-1} = [\mathbf{w}_1, \ldots, \mathbf{w}_K]^\top$ be the corresponding frozen global classifier head over $\mathcal{C}_{\text{old}}$. To approach the joint-learning goal in Eq.~\eqref{eq:efcil_objective}, we formulate an idealized feature-learning objective at stage $t$: training the current-task head $W_t$ on $\mathcal{D}_t$ while preserving decisions among old classes under $W_{T-1}$:
\begin{equation}
\begin{aligned}
\min_{\theta_t, W_t} \quad & \mathbb{E}_{(\mathbf{x}, y) \sim \mathcal{D}_t} \left[ \mathcal{L}_{\text{ce}}\left( W_t f_{\theta_t}(\mathbf{x}), y \right) \right] \\
\text{s.t.} \quad & \argmax_{c \in \mathcal{C}_{\text{old}}} \mathbf{w}_c^\top f_{\theta_t}(\mathbf{x}) = \argmax_{c \in \mathcal{C}_{\text{old}}} \mathbf{w}_c^\top f_{\theta_{t-1}}(\mathbf{x}).
\end{aligned}
\label{eq:decision_constrained}
\end{equation}
Equation~\eqref{eq:decision_constrained} translates intractable historical risk minimization into a classification-decision invariance constraint anchored by the old global head $W_{T-1}$.

\subsection{Derivation of Relative-Logit Regularization}
\label{sec:logit_derivation}

For the discontinuous $\argmax$ constraint in Eq.~\eqref{eq:decision_constrained}, we construct a continuous penalty based on a sufficient condition for decision preservation.

For any input $\mathbf{x}$, the model's relative preference between two old classes $i, j \in \mathcal{C}_{\text{old}}$ under $W_{T-1}$ is governed by their pairwise logit margin:
\begin{equation}
m_{ij}^{(s)}(\mathbf{x}) = (\mathbf{w}_i - \mathbf{w}_j)^\top f_{\theta_s}(\mathbf{x}), \qquad s \in \{t-1, t\}.
\label{eq:pairwise_margin}
\end{equation}
Preserving the $\argmax$ prediction requires that the margin between the winning class and every other old class does not flip sign. To retain the full discriminative geometry among old classes and obtain a smooth differentiable objective, we tighten sign preservation into a sufficient condition: keeping all pairwise logit differences across old classes invariant. Denoting the feature drift by $\delta f(\mathbf{x}) = f_{\theta_t}(\mathbf{x}) - f_{\theta_{t-1}}(\mathbf{x})$, this condition takes the form
\begin{equation}
\begin{aligned}
m_{ij}^{(t)}(\mathbf{x}) &= m_{ij}^{(t-1)}(\mathbf{x}) \\
&\iff (\mathbf{w}_i - \mathbf{w}_j)^\top \delta f(\mathbf{x}) = 0, \quad \forall\, i, j \in \mathcal{C}_{\text{old}}.
\end{aligned}
\label{eq:margin_invariance}
\end{equation}

Penalizing the sum of squared margin drifts over all $K(K-1)/2$ old-class pairs yields a quadratic soft constraint. Let $\delta \ell_i(\mathbf{x}) = \mathbf{w}_i^\top \delta f(\mathbf{x})$ denote the $i$-th old-logit drift and $\overline{\delta \ell}(\mathbf{x}) = \frac{1}{K}\sum_{i=1}^{K} \delta \ell_i(\mathbf{x})$ denote the mean old-logit drift. By algebraic identity, the pairwise margin penalty reduces to a centered relative-logit norm:
\begin{equation}
\begin{aligned}
\mathcal{L}_{\text{logit}}(\mathbf{x})
&= \frac{1}{K} \sum_{1 \le i < j \le K} \left( (\mathbf{w}_i - \mathbf{w}_j)^\top \delta f(\mathbf{x}) \right)^2 \\
&= \sum_{i=1}^{K} \left( \delta \ell_i(\mathbf{x}) - \overline{\delta \ell}(\mathbf{x}) \right)^2
= \left\| W_c \delta f(\mathbf{x}) \right\|_2^2,
\end{aligned}
\label{eq:logit_penalty}
\end{equation}
where $W_c = \left(I_K - \frac{1}{K}\mathbf{1}\mathbf{1}^\top\right) W_{T-1} \in \mathbb{R}^{K \times d}$ is the row-centered old classifier head.

This reduction shows that preserving all $\mathcal{O}(K^2)$ pairwise margins simplifies to an $\ell_2$ penalty on centered relative logits. Because Eq.~\eqref{eq:logit_penalty} permits all old-class logits to share an arbitrary scalar translation $a(\mathbf{x})$, namely $W_{T-1} f_{\theta_t}(\mathbf{x}) = W_{T-1} f_{\theta_{t-1}}(\mathbf{x}) + a(\mathbf{x})\mathbf{1}$, achieving $\mathcal{L}_{\text{logit}}(\mathbf{x}) = 0$ strictly preserves not only the $\argmax$ classification decision and relative class rankings, but also the exact conditional probability distribution $\operatorname{softmax}(W_{T-1} f_{\theta_t}(\mathbf{x}))$ among old classes.

\subsection{Spectral Whitening and Feature Training}
\label{sec:whitening}

To examine how $\mathcal{L}_{\text{logit}}(\mathbf{x})$ constrains feature drift in the representation space, we perform the compact singular value decomposition (SVD) on the centered old classifier head $W_c$:
\begin{equation}
W_c = U S V^\top, \qquad S = \operatorname{diag}(\sigma_1, \ldots, \sigma_r),
\label{eq:classifier_svd}
\end{equation}
where $r = \operatorname{rank}(W_c) \le \min(d, K-1)$. Substituting Eq.~\eqref{eq:classifier_svd} into Eq.~\eqref{eq:logit_penalty} yields
\begin{equation}
\begin{aligned}
\mathcal{L}_{\text{logit}}(\mathbf{x})
&= \left\| U S V^\top \delta f(\mathbf{x}) \right\|_2^2 = \left\| S V^\top \delta f(\mathbf{x}) \right\|_2^2 \\
&= \sum_{k=1}^{r} \sigma_k^2 \left( \mathbf{v}_k^\top \delta f(\mathbf{x}) \right)^2.
\end{aligned}
\label{eq:spectral_penalty}
\end{equation}
Equation~\eqref{eq:spectral_penalty} reveals mathematically that although $\mathcal{L}_{\text{logit}}(\mathbf{x})$ eliminates the redundant shared logit offset, it couples the regularization weight along each discriminative direction $\mathbf{v}_k$ to the squared singular value $\sigma_k^2$ of $W_c$. As our empirical analysis in Sec.~\ref{sec:subspace_analysis} verifies, protecting the discriminative basis $V$ identified by $W_c$ is far more effective than protecting a random orthonormal basis of equal rank $r$, and applying spectral whitening across $V$ outperforms weighting directions by $\sigma_k^2$ when controlling for average regularization scale. Motivated by these observations, we apply the spectral whitening transform $S^{-1} U^\top$ to the relative-logit drift to strip away $\sigma_k^2$ and constrain all active discriminative directions uniformly:
\begin{equation}
\mathcal{L}_{\text{dis}}(\mathbf{x}) = \left\| S^{-1} U^\top W_c \delta f(\mathbf{x}) \right\|_2^2 = \left\| V^\top \delta f(\mathbf{x}) \right\|_2^2.
\label{eq:discriminative_penalty}
\end{equation}
Because $W_{T-1}$ is frozen during stage $t$, the whitened discriminative basis $V$ is extracted once before task training, and evaluating Eq.~\eqref{eq:discriminative_penalty} requires only $\mathcal{O}(dr)$ operations per sample.

We add a mild global drift penalty to promote overall feature-space smoothness. Highlighting the contribution of $\mathcal{L}_{\mathrm{dis}}$, we write the stage-$t$ feature training objective as (with $\delta f(\mathbf{x}) = f_{\theta_t}(\mathbf{x}) - f_{\theta_{t-1}}(\mathbf{x})$):
\begin{equation}
\begin{aligned}
&\mathcal{L}_{\text{train}}(\theta_t, W_t) = \mathbb{E}_{(\mathbf{x}, y) \sim \mathcal{D}_t} \Big[ \mathcal{L}_{\text{ce}}\left( W_t f_{\theta_t}(\mathbf{x}), y \right) \\
&+ \lambda_1 \left\| V^\top \delta f(\mathbf{x}) \right\|_2^2 + \lambda_2 \left\| \left(I_d - V V^\top\right) \delta f(\mathbf{x}) \right\|_2^2 \Big],
\end{aligned}
\label{eq:training_objective}
\end{equation}
where $\lambda_1, \lambda_2 > 0$ are balancing hyperparameters. The regularization equals $\lambda_2\|\delta f(\mathbf{x})\|_2^2 + (\lambda_1-\lambda_2)\|V^\top\delta f(\mathbf{x})\|_2^2$, combining global drift control with direction-specific weighting and requiring only one low-rank projection.

\begin{figure*}[t]
\centering
\includegraphics[width=\linewidth]{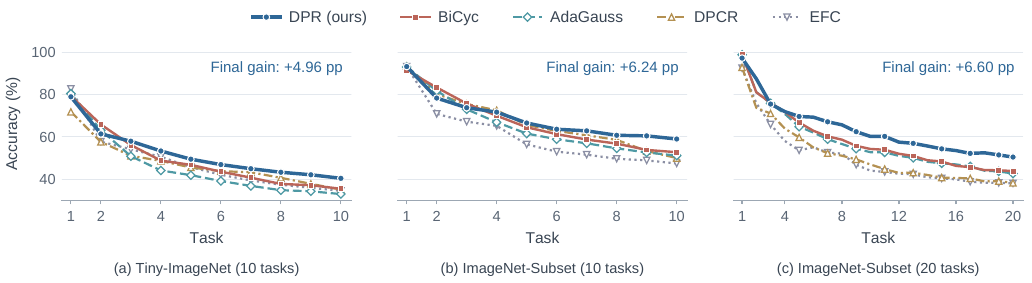}
\caption{Accuracy over all seen classes after each task on (a) Tiny-ImageNet, 10 tasks; (b) ImageNet-Subset, 10 tasks; and (c) ImageNet-Subset, 20 tasks. Final gains are in pp relative to the strongest displayed baseline.}
\label{fig:task_curves}
\end{figure*}

\subsection{Affine Statistics Compensation and Global Classifier Calibration}
\label{sec:affine_compensation}

Because \method derives the discriminative basis $V$ directly from the frozen old classifier head, sustaining its effectiveness across tasks relies on a well-calibrated global classifier $W_T$ over all seen classes $\mathcal{C}_{1:t}$ at the end of each stage. However, Eq.~\eqref{eq:training_objective} trains only the local head $W_t$, and the update from $f_{\theta_{t-1}}$ to $f_{\theta_t}$ causes stored old-class distributions to undergo location and covariance drift, preventing direct alignment with new-class features. We therefore introduce a closed-form affine transformation to compensate for old-class statistics drift and calibrate $W_T$ for both unified inference and next-stage decision regularization.

\paragraph{Affine drift estimation and statistics compensation.}
Once feature training converges, we freeze both $f_{\theta_{t-1}}$ and $f_{\theta_t}$, and extract paired features $Z = [f_{\theta_{t-1}}(\mathbf{x}_1), \ldots, f_{\theta_{t-1}}(\mathbf{x}_n)] \in \mathbb{R}^{d \times n}$ and $Z' = [f_{\theta_t}(\mathbf{x}_1), \ldots, f_{\theta_t}(\mathbf{x}_n)] \in \mathbb{R}^{d \times n}$ over the $n$ samples of $\mathcal{D}_t$. We fit a shared affine transformation $\mathbf{z}' \approx A\mathbf{z} + \mathbf{b}$ to the cross-stage feature drift via ridge regression with an identity prior:
\begin{equation}
\min_{A, \mathbf{b}} \left\| Z' - \left( A Z + \mathbf{b}\mathbf{1}_n^\top \right) \right\|_F^2 + \rho \left\| A - I_d \right\|_F^2,
\label{eq:affine_fit}
\end{equation}
where $\rho > 0$ shrinks weakly supported directions toward the identity map. Let $\bar{\mathbf{z}}, \bar{\mathbf{z}}' \in \mathbb{R}^d$ denote the sample means and $\widetilde{Z}, \widetilde{Z}'$ denote the centered feature matrices; Eq.~\eqref{eq:affine_fit} admits the closed-form solution
\begin{equation}
A = \left( \widetilde{Z}' \widetilde{Z}^\top + \rho I_d \right) \left( \widetilde{Z} \widetilde{Z}^\top + \rho I_d \right)^{-1}, \mathbf{b} = \bar{\mathbf{z}}' - A\bar{\mathbf{z}}.
\label{eq:affine_solution}
\end{equation}
For each old class $c \in \mathcal{C}_{\text{old}}$, we maintain only its first- and second-order statistics $(\boldsymbol{\mu}_c, \Sigma_c)$ in memory and transport them to the updated feature space:
\begin{equation}
\widehat{\boldsymbol{\mu}}_c = A\boldsymbol{\mu}_c + \mathbf{b}, \qquad \widehat{\Sigma}_c = A \Sigma_c A^\top, \qquad \forall\, c \in \mathcal{C}_{\text{old}}.
\label{eq:moment_transport}
\end{equation}
The means and covariances of current-task classes $c \in \mathcal{C}_t$ are computed directly from the cached features $Z'$.

\paragraph{Global classifier calibration.}
With $f_{\theta_t}$ frozen, we sample synthetic old-class features from Gaussian distributions $\mathcal{N}(\widehat{\boldsymbol{\mu}}_c, \widehat{\Sigma}_c)$ in the feature space of $f_{\theta_t}$ and jointly train the global classifier head $W_T \in \mathbb{R}^{|\mathcal{C}_{1:t}| \times d}$ alongside real current-task features:
\begin{equation}
\begin{aligned}
\mathcal{L}_{\text{cal}}(W_T) ={}& \mathbb{E}_{(\mathbf{x}, y) \sim \mathcal{D}_t} \left[ \mathcal{L}_{\text{ce}}\left( W_T f_{\theta_t}(\mathbf{x}), y \right) \right] \\
&+ \mathbb{E}_{c \in \mathcal{C}_{\text{old}}} \mathbb{E}_{\widehat{\mathbf{z}} \sim \mathcal{N}(\widehat{\boldsymbol{\mu}}_c, \widehat{\Sigma}_c)} \left[ \mathcal{L}_{\text{ce}}\left( W_T \widehat{\mathbf{z}}, c \right) \right].
\end{aligned}
\label{eq:classifier_calibration}
\end{equation}
Algorithm~\ref{alg:incremental_task} summarizes each incremental task.
\begin{algorithm}[t]
\caption{\method at task $t>1$}
\label{alg:incremental_task}
\small
\begin{algorithmic}[1]
\Require $\mathcal{D}_t$, $f_{\theta_{t-1}}$, $W_{T-1}$, old-class moments
\State Extract $V$ from centered $W_{T-1}$ via spectral whitening (Eqs.~\eqref{eq:classifier_svd}, \eqref{eq:discriminative_penalty}).
\State Optimize $(\theta_t,W_t)$ using $\mathcal{L}_{\text{train}}$ (Eq.~\eqref{eq:training_objective}).
\State Solve $(A,\mathbf{b})$ (Eq.~\eqref{eq:affine_solution}); transport old-class moments (Eq.~\eqref{eq:moment_transport}) and compute new-class moments.
\State Freeze $f_{\theta_t}$; calibrate $W_T$ using $\mathcal{L}_{\text{cal}}$ (Eq.~\eqref{eq:classifier_calibration}).
\Ensure $f_{\theta_t}$, $W_T$, and updated moments for task $t+1$
\Statex \textbf{Inference:} $\hat{y}=\argmax_{c\in\mathcal{C}_{1:t}}(W_T f_{\theta_t}(\mathbf{x}))_c$.
\end{algorithmic}
\end{algorithm}

\section{Experiments}
\label{sec:exp}

\begin{table*}[t]
\centering
\caption{Final accuracy (Acc) and average anytime accuracy (AAA), in percent. Higher is better. \textbf{Bold}: best; \underline{underline}: second best. Values are mean $\pm$ standard deviation.}
\label{tab:main_results_t10}
\footnotesize
\renewcommand{\arraystretch}{0.92}
\setlength{\tabcolsep}{7.5pt}
\begin{tabular}{lccccccc}
\toprule
\multirow{2}{*}{Method} & \multirow{2}{*}{Venue} & \multicolumn{2}{c}{CIFAR-100 ($T=10$)} & \multicolumn{2}{c}{Tiny-ImageNet ($T=10$)} & \multicolumn{2}{c}{ImageNet-Subset ($T=10$)} \\
\cmidrule(lr){3-4} \cmidrule(lr){5-6} \cmidrule(lr){7-8}
 & & Acc $\uparrow$ & AAA $\uparrow$ & Acc $\uparrow$ & AAA $\uparrow$ & Acc $\uparrow$ & AAA $\uparrow$ \\
\midrule
EWC~\cite{kirkpatrick2017overcoming} & PNAS'17 & $30.90 \pm 1.9$ & $50.40 \pm 1.7$ & $18.50 \pm 1.8$ & $34.30 \pm 2.3$ & $25.10 \pm 2.8$ & $40.60 \pm 3.3$ \\
LwF~\cite{li2018learning} & ECCV'16 & $31.90 \pm 1.1$ & $51.80 \pm 1.5$ & $27.10 \pm 1.5$ & $39.60 \pm 2.0$ & $33.40 \pm 2.2$ & $51.50 \pm 1.6$ \\
PASS~\cite{zhu2021prototype} & CVPR'21 & $30.80 \pm 1.2$ & $48.30 \pm 1.1$ & $24.50 \pm 0.6$ & $39.50 \pm 1.0$ & $26.40 \pm 1.3$ & $45.70 \pm 0.2$ \\
FeTrIL~\cite{petit2023fetril} & WACV'23 & $34.90 \pm 0.5$ & $51.20 \pm 1.1$ & $31.00 \pm 0.9$ & $45.30 \pm 1.8$ & $36.20 \pm 1.2$ & $52.60 \pm 0.6$ \\
SSRE~\cite{zhu2022self} & CVPR'22 & $31.65 \pm 1.1$ & $47.26 \pm 1.0$ & $24.15 \pm 0.7$ & $38.82 \pm 0.9$ & $26.65 \pm 1.2$ & $43.76 \pm 0.3$ \\
EFC~\cite{magistri2024elastic} & ICLR'24 & $43.50 \pm 0.8$ & $58.10 \pm 1.2$ & $34.50 \pm 1.1$ & $47.90 \pm 1.5$ & $50.90 \pm 1.1$ & $61.30 \pm 1.2$ \\
ADC~\cite{goswami2024resurrecting} & CVPR'24 & $46.50 \pm 1.2$ & $61.40 \pm 1.6$ & $32.30 \pm 1.5$ & $43.00 \pm 1.9$ & $38.30 \pm 1.2$ & $55.50 \pm 1.5$ \\
LDC~\cite{gomez2024exemplar} & ECCV'24 & $45.40 \pm 1.6$ & $59.50 \pm 1.9$ & $34.20 \pm 1.1$ & $46.80 \pm 1.6$ & $51.40 \pm 1.2$ & \underline{$69.40 \pm 0.6$} \\
AdaGauss~\cite{rypesc2024task} & NeurIPS'24 & $46.80 \pm 1.2$ & $60.90 \pm 1.0$ & $32.90 \pm 0.9$ & $45.80 \pm 1.3$ & $51.10 \pm 1.2$ & $65.00 \pm 1.4$ \\
DPCR~\cite{he2025semantic} & ICML'25 & $50.20 \pm 0.7$ & $62.80 \pm 1.1$ & $33.90 \pm 1.8$ & $46.90 \pm 0.9$ & $49.90 \pm 0.8$ & $64.80 \pm 1.1$ \\
BiCyc~\cite{xu2026two} & ICLR'26 & \underline{$50.60 \pm 0.9$} & \underline{$63.20 \pm 1.3$} & \underline{$35.40 \pm 0.8$} & \underline{$49.10 \pm 1.4$} & \underline{$52.70 \pm 0.9$} & $66.80 \pm 1.4$ \\
\midrule
\textbf{\method (ours)} & --- & \textbf{51.31} $\pm$ 0.8 & \textbf{64.04} $\pm$ 1.2 & \textbf{40.36} $\pm$ 0.9 & \textbf{51.57} $\pm$ 1.4 & \textbf{58.94} $\pm$ 1.0 & \textbf{69.43} $\pm$ 1.3 \\
\bottomrule
\end{tabular}
\end{table*}

\begin{table*}[t]
\centering
\caption{Final accuracy (Acc) and average anytime accuracy (AAA), in percent. Higher is better. \textbf{Bold}: best; \underline{underline}: second best. Values are mean $\pm$ standard deviation.}
\label{tab:main_results_t20}
\footnotesize
\renewcommand{\arraystretch}{0.92}
\setlength{\tabcolsep}{7.5pt}
\begin{tabular}{lccccccc}
\toprule
\multirow{2}{*}{Method} & \multirow{2}{*}{Venue} & \multicolumn{2}{c}{CIFAR-100 ($T=20$)} & \multicolumn{2}{c}{Tiny-ImageNet ($T=20$)} & \multicolumn{2}{c}{ImageNet-Subset ($T=20$)} \\
\cmidrule(lr){3-4} \cmidrule(lr){5-6} \cmidrule(lr){7-8}
 & & Acc $\uparrow$ & AAA $\uparrow$ & Acc $\uparrow$ & AAA $\uparrow$ & Acc $\uparrow$ & AAA $\uparrow$ \\
\midrule
EWC~\cite{kirkpatrick2017overcoming} & PNAS'17 & $17.00 \pm 1.6$ & $34.20 \pm 2.1$ & $11.30 \pm 1.9$ & $26.80 \pm 2.5$ & $13.70 \pm 2.1$ & $29.20 \pm 2.5$ \\
LwF~\cite{li2018learning} & ECCV'16 & $17.60 \pm 1.2$ & $39.20 \pm 1.7$ & $15.20 \pm 1.6$ & $31.50 \pm 2.1$ & $18.60 \pm 1.6$ & $41.30 \pm 1.9$ \\
PASS~\cite{zhu2021prototype} & CVPR'21 & $17.60 \pm 0.8$ & $31.10 \pm 1.3$ & $18.50 \pm 1.4$ & $30.40 \pm 1.9$ & $14.40 \pm 1.2$ & $31.70 \pm 0.4$ \\
FeTrIL~\cite{petit2023fetril} & WACV'23 & $23.30 \pm 1.4$ & $37.90 \pm 1.2$ & $25.90 \pm 1.2$ & $39.90 \pm 1.2$ & $26.60 \pm 1.5$ & $42.40 \pm 2.1$ \\
SSRE~\cite{zhu2022self} & CVPR'22 & $18.75 \pm 0.9$ & $32.45 \pm 1.2$ & $18.55 \pm 1.3$ & $30.62 \pm 1.8$ & $17.45 \pm 1.1$ & $31.15 \pm 0.5$ \\
EFC~\cite{magistri2024elastic} & ICLR'24 & $32.40 \pm 0.9$ & $47.00 \pm 1.3$ & $28.40 \pm 1.2$ & $42.10 \pm 1.6$ & $38.60 \pm 1.2$ & $50.50 \pm 1.5$ \\
ADC~\cite{goswami2024resurrecting} & CVPR'24 & $35.10 \pm 1.4$ & $51.70 \pm 1.8$ & $18.10 \pm 1.6$ & $36.00 \pm 2.1$ & $25.10 \pm 1.3$ & $43.40 \pm 1.7$ \\
LDC~\cite{gomez2024exemplar} & ECCV'24 & $35.50 \pm 1.9$ & $51.90 \pm 2.3$ & $24.90 \pm 2.2$ & $38.20 \pm 2.7$ & $28.50 \pm 1.7$ & $46.50 \pm 2.7$ \\
AdaGauss~\cite{rypesc2024task} & NeurIPS'24 & $37.90 \pm 1.0$ & $54.40 \pm 0.8$ & $27.50 \pm 1.2$ & $39.50 \pm 1.1$ & $42.60 \pm 1.6$ & $57.40 \pm 1.9$ \\
DPCR~\cite{he2025semantic} & ICML'25 & $39.80 \pm 1.2$ & $54.80 \pm 0.9$ & $25.60 \pm 0.7$ & $39.30 \pm 0.6$ & $37.30 \pm 1.6$ & $54.70 \pm 0.7$ \\
BiCyc~\cite{xu2026two} & ICLR'26 & \textbf{41.50} $\pm$ 1.1 & \textbf{56.50} $\pm$ 1.3 & \underline{30.20} $\pm$ 1.1 & \underline{44.20} $\pm$ 1.3 & \underline{43.80} $\pm$ 1.4 & \underline{58.20} $\pm$ 1.8 \\
\midrule
\textbf{\method (ours)} & --- & \underline{41.22} $\pm$ 1.1 & \underline{54.82} $\pm$ 1.5 & \textbf{35.25} $\pm$ 1.1 & \textbf{47.37} $\pm$ 1.4 & \textbf{50.40} $\pm$ 1.3 & \textbf{63.39} $\pm$ 1.6 \\
\bottomrule
\end{tabular}
\end{table*}

In this section, we first present the experimental setup (Sec.~\ref{sec:settings}), compare \method with state-of-the-art EFCIL methods on standard and large-scale benchmarks (Sec.~\ref{sec:main_results}), and then analyze component contributions, decision preservation and plasticity, the discriminative basis and spectral whitening, and runtime (Secs.~\ref{sec:ablation} to \ref{sec:efficiency}).

\subsection{Experimental Settings}
\label{sec:settings}

\paragraph{Datasets and metrics.}
Our main evaluation uses CIFAR-100~\cite{krizhevsky2009learning}, Tiny-ImageNet~\cite{wu2017tiny}, and ImageNet-Subset~\cite{deng2009imagenet}, and we further evaluate scalability on full ImageNet-1K~\cite{deng2009imagenet}. Following the cold-start EFCIL protocol~\cite{magistri2024elastic}, classes are split evenly into $T=10$ or $T=20$ sequential tasks without a pre-trained or large initial task. CIFAR-100 and ImageNet-Subset contain 100 classes (10 or 5 classes per task), Tiny-ImageNet contains 200 classes (20 or 10 classes per task), and ImageNet-1K contains 1,000 classes (100 or 50 classes per task). CIFAR-100 and Tiny-ImageNet provide 500 training images per class and 100 or 50 test images per class, respectively. ImageNet-Subset uses the standard 100-class split~\cite{douillard2020podnet} with approximately 1,300 training and 50 validation images per class. Let $a_t$ denote classification accuracy over all classes seen through task $t$ ($\mathcal{C}_{1:t}$), evaluated after global classifier calibration without task identifiers. We report final accuracy $\mathrm{Acc} = a_T$ ($\uparrow$) and average anytime accuracy $\mathrm{AAA} = \frac{1}{T}\sum_{t=1}^T a_t$ ($\uparrow$) in percent as mean $\pm$ standard deviation over five runs.

\paragraph{Competing methods and implementation details.}
We compare \method against eleven representative baselines: EWC~\cite{kirkpatrick2017overcoming}, LwF~\cite{li2018learning}, PASS~\cite{zhu2021prototype}, FeTrIL~\cite{petit2023fetril}, SSRE~\cite{zhu2022self}, EFC~\cite{magistri2024elastic}, ADC~\cite{goswami2024resurrecting}, LDC~\cite{gomez2024exemplar}, AdaGauss~\cite{rypesc2024task}, DPCR~\cite{he2025semantic}, and BiCyc~\cite{xu2026two}. Following standard cold-start EFCIL settings~\cite{magistri2024elastic,magistri2025efc}, all methods use a ResNet-18 backbone ($d=512$) trained from scratch. We optimize incremental representation learning (Eq.~\eqref{eq:training_objective}) using Adam with weight decay $2\times 10^{-4}$, a fixed learning rate, 100 epochs per task, and batch size 64. The learning rate is $10^{-4}$ for both the backbone and local head on CIFAR-100 and Tiny-ImageNet, and $10^{-5}$ for the backbone and $10^{-4}$ for the head on ImageNet-Subset. Across all datasets, we fix the regularization coefficients to $\lambda_1 = 0.5$ and $\lambda_2 = 0.01$. Closed-form affine compensation (Eq.~\eqref{eq:affine_solution}) uses ridge parameter $\rho = 10^{-3}$, and global classifier calibration (Eq.~\eqref{eq:classifier_calibration}) trains the bias-free linear head $W_T$ for 50 epochs (batch size 256) using SGD with fixed learning rate $10^{-3}$ on sampled Gaussian features and cached current-task features. All experiments are conducted on a single NVIDIA GeForce RTX 3090 Ti GPU.

\subsection{Main Results}
\label{sec:main_results}

\method achieves the highest final accuracy and AAA on all three datasets with 10 tasks, and on Tiny-ImageNet and ImageNet-Subset with 20 tasks (Tables~\ref{tab:main_results_t10} and \ref{tab:main_results_t20}). The final-accuracy gains over the strongest baseline are largest on Tiny-ImageNet and ImageNet-Subset, ranging from 4.96 to 6.60 pp. With 20 tasks, AAA also improves by 3.17 pp on Tiny-ImageNet and 5.19 pp on ImageNet-Subset, showing that the benefit extends to average performance over the learning sequence. The task-wise curves in Fig.~\ref{fig:task_curves} compare \method with EFC, DPCR, AdaGauss, and BiCyc, showing its advantage through the later stages of learning.

\begin{table}[t]
\centering
\caption{\textbf{Large-scale evaluation on ImageNet-1K.}
Final accuracy ($\mathrm{Acc}$, $\%$) and average anytime accuracy ($\mathrm{AAA}$, $\%$) under 10- and 20-task cold-start splits.}
\label{tab:imagenet1k}
\footnotesize
\renewcommand{\arraystretch}{0.92}
\setlength{\tabcolsep}{6pt}
\begin{tabular}{lcccc}
\toprule
\multirow{2}{*}{Method} & \multicolumn{2}{c}{$T=10$ (100 cls/task)} & \multicolumn{2}{c}{$T=20$ (50 cls/task)} \\
\cmidrule(lr){2-3} \cmidrule(lr){4-5}
 & Acc $\uparrow$ & AAA $\uparrow$ & Acc $\uparrow$ & AAA $\uparrow$ \\
\midrule
BiCyc~\cite{xu2026two} & 38.48 & 52.37 & 32.74 & 49.12 \\
\textbf{\method (ours)} & \textbf{42.31} & \textbf{56.30} & \textbf{36.17} & \textbf{50.16} \\
\bottomrule
\end{tabular}
\end{table}

\paragraph{Large-scale scenarios.}
\method also outperforms BiCyc on full ImageNet-1K under both task splits (Table~\ref{tab:imagenet1k}), using the same incremental-stage configuration as ImageNet-Subset. Final accuracy improves by 3.83 and 3.43 pp for 10 and 20 tasks, respectively, while AAA improves by 3.93 and 1.04 pp. These gains show that \method remains effective as the number of classes grows to 1,000.

\begin{table*}[t]
\centering
\caption{\textbf{Component ablations and coefficient sensitivity on CIFAR-100.}
(a) Checkmarks indicate enabled components; $\mathrm{Acc}$ and $\mathrm{AAA}$ are percentages.
(b, c) Final accuracy ($\%$) across coefficient pairs. Boxes mark $(\lambda_1, \lambda_2) = (0.5, 0.01)$; darker cells indicate higher accuracy within each panel.}
\label{tab:ablation_sensitivity}
\small
\renewcommand{\arraystretch}{1.06}
\setlength{\tabcolsep}{3pt}
\setlength{\fboxsep}{1.3pt}
\setlength{\fboxrule}{0.6pt}
\captionsetup[subtable]{font=small,labelfont=bf,textfont=bf,skip=4pt}
\begin{subtable}[t]{0.35\textwidth}
\centering
\caption{Component ablations ($T=10$).}
\begin{tabularx}{\linewidth}{*{3}{>{\centering\arraybackslash}X}*{2}{>{\raggedleft\arraybackslash}X}}
\toprule
\multicolumn{3}{c}{\textbf{Enabled components}} & \multicolumn{2}{c}{\textbf{Accuracy (\%)}} \\
\cmidrule(lr){1-3}\cmidrule(l){4-5}
$\mathcal{L}_{\text{dis}}$ & $\delta f$ & Affine & Acc $\uparrow$ & AAA $\uparrow$ \\
\midrule
\cmark & \textcolor{black!40}{\xmark} & \cmark & 49.27 & 63.57 \\
\textcolor{black!40}{\xmark} & \cmark & \cmark & 41.40 & 59.54 \\
\cmark & \cmark & \textcolor{black!40}{\xmark} & 46.95 & 61.93 \\
\midrule
\rowcolor{black!5}
\cmark & \cmark & \cmark & \textbf{51.31} & \textbf{64.04} \\
\bottomrule
\end{tabularx}
\end{subtable}%
\hfill
\begin{subtable}[t]{0.30\textwidth}
\centering
\caption{Sensitivity on $T=10$ ($\mathrm{Acc}$, $\%$).}
\begin{tabularx}{\linewidth}{>{\centering\arraybackslash}p{16pt}*{3}{>{\raggedleft\arraybackslash}X}}
\toprule
 & \multicolumn{3}{c}{$\lambda_2$} \\
\cmidrule(l){2-4}
$\lambda_1$ & \multicolumn{1}{c}{0.01} & \multicolumn{1}{c}{0.02} & \multicolumn{1}{c}{0.03} \\
\midrule
\rule[-4pt]{0pt}{17pt}0.4 & \cellcolor{bpOurs!36}51.15 & \cellcolor{bpOurs!12}50.44 & \cellcolor{bpOurs!8}50.33 \\
\rule[-4pt]{0pt}{17pt}0.5 & \cellcolor{bpOurs!42}\fcolorbox{bpOurs}{bpOurs!42}{\textbf{51.31}}\hspace*{-1.9pt} & \cellcolor{bpOurs!33}51.04 & \cellcolor{bpOurs!12}50.45 \\
\rule[-4pt]{0pt}{17pt}0.6 & \cellcolor{bpOurs!36}51.13 & \cellcolor{bpOurs!31}51.00 & \cellcolor{bpOurs!9}50.36 \\
\bottomrule
\end{tabularx}
\end{subtable}%
\hfill
\begin{subtable}[t]{0.30\textwidth}
\centering
\caption{Sensitivity on $T=20$ ($\mathrm{Acc}$, $\%$).}
\begin{tabularx}{\linewidth}{>{\centering\arraybackslash}p{16pt}*{3}{>{\raggedleft\arraybackslash}X}}
\toprule
 & \multicolumn{3}{c}{$\lambda_2$} \\
\cmidrule(l){2-4}
$\lambda_1$ & \multicolumn{1}{c}{0.01} & \multicolumn{1}{c}{0.02} & \multicolumn{1}{c}{0.03} \\
\midrule
\rule[-4pt]{0pt}{17pt}0.4 & \cellcolor{bpOurs!42}\textbf{41.53} & \cellcolor{bpOurs!30}41.11 & \cellcolor{bpOurs!8}40.36 \\
\rule[-4pt]{0pt}{17pt}0.5 & \cellcolor{bpOurs!33}\fcolorbox{bpOurs}{bpOurs!33}{41.22}\hspace*{-1.9pt} & \cellcolor{bpOurs!28}41.04 & \cellcolor{bpOurs!21}40.82 \\
\rule[-4pt]{0pt}{17pt}0.6 & \cellcolor{bpOurs!38}41.40 & \cellcolor{bpOurs!25}40.96 & \cellcolor{bpOurs!19}40.74 \\
\bottomrule
\end{tabularx}
\end{subtable}
\end{table*}

\subsection{Component Ablations}
\label{sec:ablation}

\paragraph{Regularization terms.}
On CIFAR-100 with 10 tasks, we remove each component in turn (Table~\ref{tab:ablation_sensitivity}a). Removing the whitened decision-preserving penalty $\mathcal{L}_{\text{dis}}$ reduces final accuracy by 9.91 pp and AAA by 4.50 pp, showing its central role in preserving performance. Adding the global feature-drift penalty $\delta f$ to $\mathcal{L}_{\text{dis}}$ improves final accuracy by 2.04 pp and also improves AAA, providing an additional benefit by constraining feature drift. We further analyze sensitivity to $\lambda_1 \in \{0.4, 0.5, 0.6\}$ and $\lambda_2 \in \{0.01, 0.02, 0.03\}$ (Table~\ref{tab:ablation_sensitivity}b, c). Final accuracy varies by 0.98 and 1.17 pp across the grid for 10 and 20 tasks, respectively, indicating limited sensitivity within this neighborhood.

\paragraph{Affine compensation.}
Using untransported old-class means and covariances during global classifier calibration lowers final accuracy by 4.36 pp and AAA by 2.12 pp (Table~\ref{tab:ablation_sensitivity}a). Affine transport therefore keeps the stored distributions aligned with the evolving representation.

\begin{figure*}[!t]
\centering
\includegraphics[width=\linewidth]{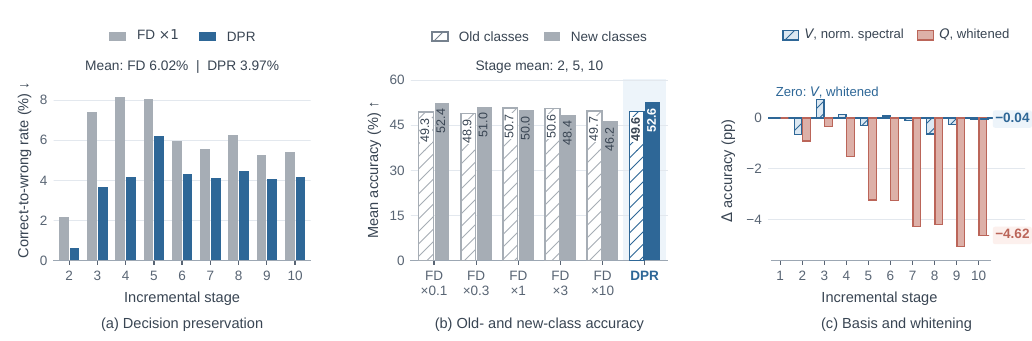}
\caption{\textbf{Protecting discriminative directions preserves decisions while allowing new-class learning.}
(a) On 10-task Tiny-ImageNet, \method (blue) reduces fixed-old-head correct-to-wrong rates at every stage relative to trace-matched feature distillation (FD, gray).
(b) After calibration, old/new class accuracies are averaged equally over stages 2, 5, and 10; \method achieves the highest old/new mean across all FD strengths.
(c) On 10-task CIFAR-100, replacing $V$ with a random basis $Q$ loses 4.62 pp; normalized spectral weighting stays within 0.74 pp of whitening.}
\label{fig:svd_ablation}
\end{figure*}

\subsection{Decision Preservation and New-Class Plasticity}
\label{sec:decision_plasticity}

We compare \method with isotropic feature distillation (FD) on 10-task Tiny-ImageNet to examine how directional protection affects old decisions and new-class learning. At each stage, both updates start from the same preceding \method model, head, and stored statistics, and apply the same affine compensation and classifier calibration. Each FD branch runs one incremental stage; subsequent stages start from the \method trajectory. FD penalizes $\mu_t(s)\|\delta f\|_2^2$, where $\mu_t(s)=s[\lambda_1r+\lambda_2(d-r)]/d$ and $s=1$ matches the trace of \method's regularization matrix.

\paragraph{Preserving correct old-class decisions.}
\method reduces decision damage at every incremental stage (Fig.~\ref{fig:svd_ablation}a). With the old head fixed and prediction restricted to old classes, we measure the fraction of previously correct old-class examples that become incorrect after the backbone update. Averaged equally over stages 2 to 10, this rate falls from $6.02\%$ for trace-matched FD to $3.97\%$ for \method, a reduction of 2.05 pp. Thus, concentrating protection on discriminative directions better preserves old-class decisions during new-class training.

\paragraph{Old/new-class tradeoff.}
\method achieves the highest mean of old- and new-class accuracy across the FD coefficient sweep (Fig.~\ref{fig:svd_ablation}b). We evaluate $s\in\{0.1,0.3,1,3,10\}$ at stages 2, 5, and 10. After calibration, predictions cover all seen classes; old- and new-class macro accuracies are averaged equally over these three stages. \method attains $49.56\%$ old-class and $52.57\%$ new-class accuracy, exceeding both metrics of the two weakest FD configurations. Relative to trace-matched FD, \method improves new-class accuracy by 2.57 pp with a 1.09 pp decrease in old-class accuracy. Its old/new mean reaches $51.06\%$, compared with $50.87\%$ for the best FD setting. Stronger isotropic penalties further reduce new-class accuracy, supporting selective protection of the old classifier's discriminative directions.

\subsection{Discriminative Basis and Spectral Whitening}
\label{sec:subspace_analysis}

\paragraph{Discriminative basis.}
To test whether the row-centered old classifier head $W_c$ identifies the directions critical for retention, we replace $V$ on 10-task CIFAR-100 with a random orthonormal basis $Q \in \mathbb{R}^{d \times r}$ of identical rank $r$ under the same whitened form, updating the complementary projector to $I_d - QQ^\top$. Replacing $V$ with $Q$ lowers accuracy at every incremental stage (Fig.~\ref{fig:svd_ablation}c), reducing final accuracy from $51.04\%$ to $46.42\%$ ($-4.62$ pp) and AAA from $63.81\%$ to $61.08\%$ ($-2.73$ pp). This supports protecting the discriminative directions identified by the row space of $W_c$.

\paragraph{Spectral whitening.}
Next, we retain the classifier-derived basis $V$ and compare spectral whitening against normalized spectral weights $\alpha_k = \sigma_k^2 / (r^{-1}\sum_{j=1}^r \sigma_j^2)$, whose mean is normalized to one so that the overall regularization scale matches $\lambda_1$. Spectral whitening achieves $51.04\%$ final accuracy and $63.81\%$ AAA, compared with $51.00\%$ and $63.72\%$ for normalized spectral weighting. Stage-wise differences remain below 0.74 pp across all ten stages (Fig.~\ref{fig:svd_ablation}c). Whitening therefore removes dependence on the classifier's singular-value scales while retaining comparable accuracy and a small final advantage, supporting Eq.~\eqref{eq:discriminative_penalty}.

\begin{figure}[t]
\centering
\includegraphics[width=\linewidth]{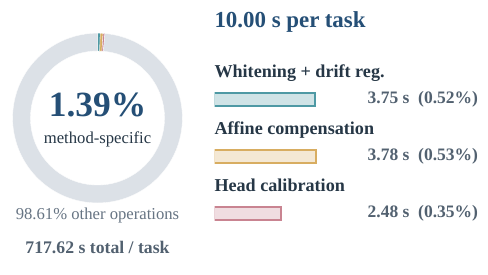}
\caption{\textbf{Method-specific operations account for only 1.39\% of runtime.}
Runtime shares computed from mean component times over Stages 2 to 10 on Tiny-ImageNet ($T=10$). Whitening and drift regularization (teal), affine compensation (gold), and classifier calibration (rose) total 10.00 s of the 717.62 s average task; grey denotes all other operations.}
\label{fig:efficiency}
\end{figure}

\subsection{Efficiency}
\label{sec:efficiency}

As shown in Fig.~\ref{fig:teaser}, \method achieves $58.94\%$ and $50.40\%$ final accuracy on ImageNet-Subset with 10 and 20 tasks, respectively. Its 10-task training takes 5.04 hours, compared with 9.40 hours for BiCyc, and is $1.05\times$, $2.12\times$, and $2.41\times$ faster than EFC, AdaGauss, and DPCR, respectively.

\paragraph{Component-wise runtime efficiency.}
Our three method-specific components account for only $1.39\%$ of per-task runtime on 10-task Tiny-ImageNet (Fig.~\ref{fig:efficiency}). Averaged over Stages 2 to 10, an incremental task takes 717.62 s: whitening and drift regularization cost 3.75 s, closed-form affine compensation 3.78 s, and global classifier calibration 2.48 s, totaling 10.00 s. Basis extraction runs once per task, affine compensation uses a closed-form solve, and calibration operates directly on features. See the supplementary material for per-task runtimes.

\begin{table}[t]
\centering
\caption{\textbf{Storage analysis of compared methods.}}
\label{tab:efficiency_storage}
\footnotesize
\renewcommand{\arraystretch}{0.92}
\setlength{\tabcolsep}{3.2pt}
\begin{tabular}{lcccc}
\toprule
Method & Prototypes & Covariances & Extra Params. & Discardable \\
\midrule
EFC~\cite{magistri2024elastic} & \cmark & \cmark & \xmark & \cmark \\
AdaGauss~\cite{rypesc2024task} & \cmark & \cmark & \cmark & \xmark \\
DPCR~\cite{he2025semantic} & \cmark & \cmark & \xmark & \cmark \\
BiCyc~\cite{xu2026two} & \cmark & \cmark & \cmark & \xmark \\
\textbf{\method (ours)} & \cmark & \cmark & \xmark & \cmark \\
\bottomrule
\end{tabular}
\end{table}

\paragraph{Storage Analysis.}
Like prior methods, \method stores per-class means and symmetric covariances $(\boldsymbol{\mu}_c, \Sigma_c)$ during training (Table~\ref{tab:efficiency_storage}). Extra parameters are counted beyond the backbone and classifier, and ``Discardable'' indicates whether stored statistics can be removed at inference. Although AdaGauss and BiCyc store lower-dimensional statistics, their projection layers introduce extra parameters, and these statistics must be retained for Mahalanobis inference.

\section{Conclusion}
\label{sec:conclusion}

We presented Decision-Preserving Regularization (\method) for cold-start exemplar-free class-incremental learning. Reformulating the intractable joint-learning objective as classification-decision invariance under the frozen old classifier reduces pairwise margin preservation to a relative-logit penalty on the row-centered classifier head. Applying spectral whitening decouples the classifier's spectral scaling and extracts an orthonormal discriminative basis directly from the classifier weights, constraining active decision directions in $\mathcal{O}(dr)$ time per sample while leaving complementary directions lightly regularized for new-class adaptation. Because anchoring decision invariance across sequential stages relies on a well-calibrated global classifier over all seen classes, we pair whitened regularization with closed-form ridge-regression affine transport of old-class Gaussian moments and feature-space classifier calibration. Together, these two components make \method simple, efficient, and effective across CIFAR-100, Tiny-ImageNet, ImageNet-Subset, and ImageNet-1K.

\paragraph{Limitations.}
Fixed $(\lambda_1, \lambda_2) = (0.5, 0.01)$ performs robustly across benchmarks, although longer task sequences mildly favor $\lambda_1 = 0.4$ (Table~\ref{tab:ablation_sensitivity}b, c). The rank $r \le \min(d, K-1)$ can reach $d$ when $K>d$ and $W_c$ has full column rank; adaptive spectral truncation is a promising direction for thousand-class learning.

\section*{LLM Usage Statement}
We used OpenAI's ChatGPT and Codex to assist with language polishing and internal manuscript review. The authors are responsible for the final content.

{
    \small
    \bibliographystyle{ieeenat_fullname}
    \bibliography{main}
}

\clearpage
\onecolumn
\appendix
\setcounter{figure}{0}
\renewcommand{\thefigure}{A\arabic{figure}}

\section{Per-Task Runtime Analysis}
\label{sec:runtime_appendix}

\begin{figure}[ht]
\centering
\includegraphics[width=0.9\linewidth]{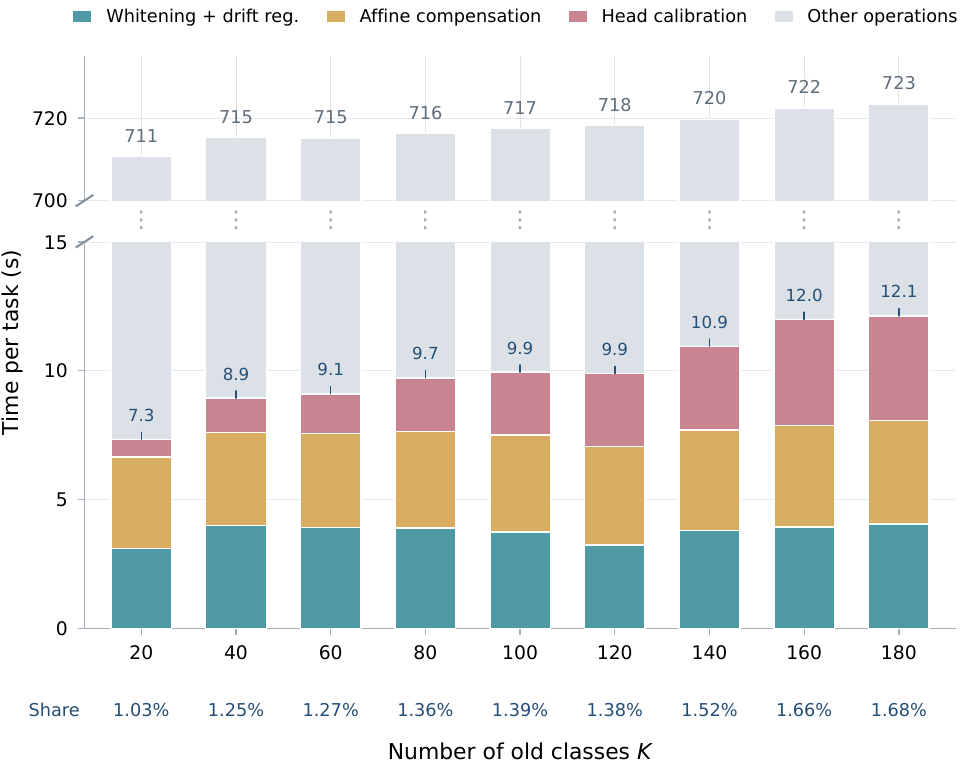}
\caption{\textbf{Method-specific runtime remains below 2\% at every incremental stage.}
Per-task wall-clock time on 10-task Tiny-ImageNet. Stacked components use the same colors as Fig.~\ref{fig:efficiency} in the main paper; grey denotes all other operations. The broken axis separates component costs from total task time. Labels show total seconds, component totals, and their shares.}
\label{fig:efficiency_per_task}
\end{figure}

Figure~\ref{fig:efficiency_per_task} expands the mean runtime breakdown in Sec.~\ref{sec:efficiency} of the main paper into individual incremental tasks on Tiny-ImageNet ($T=10$, ResNet-18, $d=512$). Across Stages 2 to 10, the three method-specific components take 7.34 to 12.13 s per task, or $1.03\%$ to $1.68\%$ of total runtime, as the number of old classes grows from 20 to 180.

The basis extraction takes 2.82 ms per task on average. Directional regularization costs 0.24 ms per batch, compared with 32.22 ms for classification forward/backward; using $\|(I_d-VV^\top)\delta f\|_2^2=\|\delta f\|_2^2-\|V^\top\delta f\|_2^2$ reduces this to 0.15 ms. Affine compensation takes 3.52 s for paired feature extraction, 5.05 ms for the ridge solve, and 0.25 s for old-class moment transport. Global classifier calibration takes 0.68 to 4.11 s as the number of seen classes increases from 40 to 200. Runtime shares use the full incremental-task time, including backbone training and the frozen teacher forward pass; the regularization cost is estimated from batch microbenchmarks.

\end{document}